\documentclass[letterpaper, 10 pt, conference]{ieeeconf}  %
\usepackage{times}
\usepackage{luatex85}

\usepackage{ifthen}

\newcommand{\fcnet}[6]{

    \begin{tikzpicture}[shorten >=1pt,->,draw=black!50]

        \tikzstyle{neuron}=[circle,fill=gray!50,minimum size=17pt] %

        \def\layersep{#1}
        \def\nodesep{#2}

        \pgfmathsetmacro{\ilw}{#3}
        \pgfmathsetmacro{\nhl}{#4}
        \pgfmathsetmacro{\hlw}{#5}
        \pgfmathsetmacro{\olw}{#6}

        \pgfmathtruncatemacro{\iof}{(\ilw*\nodesep)/2}
        \pgfmathtruncatemacro{\hof}{(\hlw*\nodesep)/2}
        \pgfmathtruncatemacro{\oof}{(\olw*\nodesep)/2}

        \foreach \name / \y in {1,...,\ilw}
            \path[yshift=\iof]
                node[neuron] (I-\name) at (0,-\y * \nodesep) {};

        \foreach \namei / \yi in {1,...,\nhl}
            \foreach \namej / \yj in {1,...,\hlw}
                \path[yshift=\hof]
                    node[neuron] (H-\namei\namej) at (\yi * \layersep,-\yj * \nodesep) {};

        \foreach \name / \y in {1,...,\olw}
            \path[yshift=\oof]
                node[neuron] (O-\name) at (\nhl*\layersep + \layersep,-\y * \nodesep) {};

        \foreach \source in {1,...,\ilw}
            \foreach \dest in {1,...,\hlw}
                \path (I-\source) edge (H-1\dest);

        \ifthenelse{\nhl > 1}{
            \pgfmathtruncatemacro{\nhlm}{\nhl - 1}
            \foreach \layer in {1,...,\nhlm}
                \pgfmathtruncatemacro{\nextlayer}{\layer + 1}
                \foreach \source in {1,...,\hlw}
                    \foreach \dest in {1,...,\hlw}
                        \path (H-\layer\source) edge (H-\nextlayer\dest);
            }{}

        \foreach \source in {1,...,\hlw}
            \foreach \dest in {1,...,\olw}
                \path (H-\nhl\source) edge (O-\dest);

    \end{tikzpicture}
}

\usepackage[bookmarks=true]{hyperref}

\usepackage{booktabs}       %
\usepackage{amsfonts,amssymb,mathtools}       %

\usepackage{subcaption}
\usepackage{multirow}

\usepackage{flushend}
\usepackage{algorithm}
\usepackage{algorithmicx}
\usepackage[noend]{algpseudocode}
\usepackage{cleveref}
\usepackage{acronym}
\usepackage{verbatim}
\usepackage{siunitx}

\crefname{appsec}{Appendix}{Appendices}

\newcommand{\etal}{{et~al.}}

\newcommand{\Lag}{\mathcal{L}} %

\newcommand{\ELeqn}{\frac{d}{dt}\left(\frac{\partial\Lag}{\partial\dot{q}}\right) - \frac{\partial\Lag}{\partial q} = F(q,\dot{q},u)} %

\newcommand\bbR{\ensuremath{\mathbb{R}}} %
\newcommand\qdot{\ensuremath{\dot{q}}} %
\newcommand\qddot{\ensuremath{\ddot{q}}} %
\newcommand\M{\ensuremath{\mathbf{M}}} %
\newcommand\C{\ensuremath{\mathbf{C}}} %
\newcommand\B{\ensuremath{\mathbf{B}}} %

\DeclarePairedDelimiter{\norm}{\lVert}{\rVert}
\newcommand{\Dtr}{\mathcal{D}_{\text{train}}}
\newcommand{\Dte}{\mathcal{D}_{\text{val}}}

\newcommand\Mnn{\ensuremath{\M_\theta(q)}}
\newcommand\Vnn{\ensuremath{V_\theta(q)}}
\newcommand\Fnn{\ensuremath{F_\theta(q,\qdot,u)}}
\newcommand\Bnn{\ensuremath{\B_\theta(q)u + \eta_{wb}\circ \qdot}}
\newcommand\Mwb{\ensuremath{\M_{wb}(q)}}
\newcommand\Vwb{\ensuremath{V_{wb}(q)}}
\newcommand\Fwb{\ensuremath{\B_{wb}u + \eta_{wb}\circ \qdot}}

\newcommand\model{\ensuremath{\mathcal{T}}}

\usepackage{todonotes}
\usepackage{soul}
\definecolor{smoothgreen}{rgb}{0.7,1,0.7}
\sethlcolor{smoothgreen}

\usepackage[utf8]{inputenc}
\RequirePackage{luatex85}
\usepackage{pgfplots}
\pgfplotsset{compat=newest}
\pgfplotsset{every axis legend/.append style={%
cells={anchor=west}}
}
\usepgfplotslibrary{polar}
\usetikzlibrary{arrows}
\tikzset{>=stealth'}

\definecolor{C1}{rgb}{0.0, 0.447, 0.741}
\definecolor{C1_light}{rgb}{0.0, 0.6032388663967612, 1.0}
\definecolor{C2}{rgb}{0.85, 0.325, 0.098}
\definecolor{C3}{rgb}{0.929, 0.694, 0.125}
\definecolor{C4}{rgb}{0.494, 0.184, 0.556}
\definecolor{C5}{rgb}{0.466, 0.674, 0.188}
\definecolor{C6}{rgb}{0.301, 0.745, 0.933}
\definecolor{C7}{rgb}{0.635, 0.078, 0.184}

\definecolor{nice-red}{HTML}{E41A1C}
\definecolor{nice-orange}{HTML}{FF7F00}
\definecolor{nice-yellow}{HTML}{FFC020}
\definecolor{nice-green}{HTML}{4DAF4A}
\definecolor{nice-blue}{HTML}{377EB8}
\definecolor{nice-nice-red}{HTML}{984EA3}
\usepgfplotslibrary{groupplots}

\usetikzlibrary{shapes.geometric, arrows}

\tikzstyle{startstop} = [rectangle, rounded corners, minimum width=2cm, minimum height=1cm,text centered, draw=black, fill=none]
\tikzstyle{arrow} = [thick,->,>=stealth]

\usetikzlibrary{backgrounds}

\usepackage{ifthen}

\algnewcommand{\LineComment}[1]{\State\(\triangleright\) #1}

\IEEEoverridecommandlockouts                              %

\title{\LARGE \bf
A General Framework for Structured Learning of Mechanical Systems
}

\author{Jayesh K. Gupta$^{1,*}$, Kunal Menda$^{1,*}$, Zachary Manchester$^{1}$, Mykel J. Kochenderfer$^{1}$%
\thanks{$^*$Authors contributed equally.}
\thanks{$^{1}$Jayesh K. Gupta, Kunal Menda, Zachary Manchester and Mykel J. Kochenderfer are at Stanford University, Stanford, CA 94305, USA
        {\tt\scriptsize \{jayeshkg,kmenda,zacmanchester,mykel\}@stanford.edu}}%
}

\renewcommand{\baselinestretch}{0.96}

\begin{document}

\maketitle
\thispagestyle{plain}
\pagestyle{plain}

\begin{abstract}
Learning accurate dynamics models is necessary for optimal, compliant control of robotic systems.
Current approaches to white-box modeling using analytic parameterizations, or black-box modeling using neural networks, can suffer from high bias or high variance. 
We address the need for a flexible, gray-box model of mechanical systems that can seamlessly incorporate prior knowledge where it is available, and train expressive function approximators where it is not. 
We propose to parameterize a mechanical system using neural networks to model its Lagrangian and the generalized forces that act on it. 
We test our method on a simulated, actuated double pendulum.
We show that our method outperforms a naive, black-box model in terms of data-efficiency, as well as performance in model-based reinforcement learning.
We also conduct a systematic study of our method's ability to incorporate available prior knowledge about the system to improve data efficiency.
\end{abstract}

\IEEEpeerreviewmaketitle

\section{Introduction}
\label{sec:introduction}

When engineering a controller for a robotic domain, we often rely on accurate models of the system we aim to control~\cite{de2012theory,ioannou2012robust}. 
One faces the perennial question of whether to seek out domain expertise, or to take a data-driven, black-box approach to constructing such a model.
The former approach would make assumptions about the system, such as its kinematic structure, inertia properties, and assumptions regarding the forces acting on the system, leaving only a few parameters for data-driven calibration~\cite{an1985estimation,Astrom1971-ia}. 
The latter approach~\cite{werbos1992handbook, raissi2018multistep, chen1990non}, on the other hand, would treat the system's equations of motion as any other function that the tools of machine learning are capable of fitting. 
That is, this approach would reduce the problem of learning the system dynamics to that of optimizing the parameters of an expressive function class, such as a neural network, in order to minimize some form of a prediction loss. 

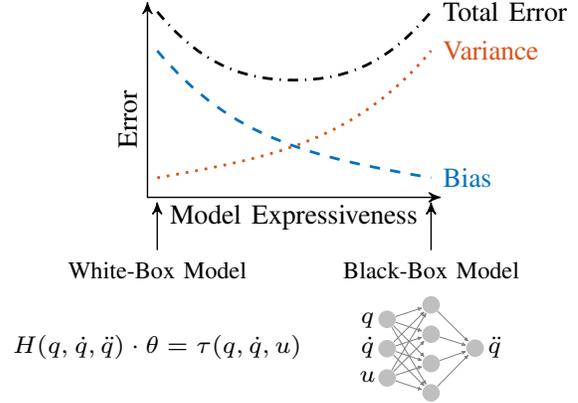
\begin{figure}
    \centering
    \scalebox{1.3}{\begin{tikzpicture}

	\draw [<->] (0,2) -- (0,0) -- (3,0);

	\draw[thick, draw=C1, dashed] (0.1,1.5) to [out=-50,in=170] (2.9,0.2);

	\draw[thick, draw=C2, dotted] (0.1,0.2) to [out=10,in=230] (2.9,1.5);

	\draw[dashdotted, thick] (0.1, 1.9) to [out=-50, in=180] (1.5, 1.2) to [out=0, in=230] (2.9,1.9);

	\node[anchor=west,text=C1] (Bias) at (2.9, 0.2) {\footnotesize Bias};

	\node[anchor=west,text=C2] (Variance) at (2.9, 1.5) {\footnotesize Variance};
	\node[anchor=west] (TotalErr) at (2.9, 1.9) {\footnotesize Total Error};

	\node (xlabel) at (1.5, -.2) {\footnotesize Model Expressiveness};
	\node[rotate=90] (ylabel) at (-0.2,0.8) {\footnotesize Error}; 

	\node[text=black] (wbmodel) at (0.1, -0.75) {\scriptsize White-Box Model};
	\draw [->] (wbmodel.north) -- (0.1,-0.05);
	\node[text=black] (bbmodel) at (2.9, -0.75) {\scriptsize Black-Box Model};
	\draw [->] (bbmodel.north) -- (2.9,-0.05);

	\node[below of=wbmodel, yshift=0.22 cm] (wbtext) {\scriptsize $H(q,\dot{q}, \ddot{q})\cdot \theta = \tau(q,\dot{q},u)$};

	\node[below of=bbmodel, yshift=0.2 cm] (bbnn) {\scalebox{0.3}{\fcnet{1.5 cm}{1cm}{3}{1}{4}{1}}};

	\node[below of=bbmodel, xshift=-0.65 cm, yshift=0.22 cm] (qdot) {\scriptsize $\dot{q}$};
	\node[above of=qdot, yshift=-0.73 cm] {\scriptsize $q$};
	\node[below of=qdot, yshift=0.7 cm] {\scriptsize $u$};
	\node[below of=bbmodel, xshift=0.65 cm, yshift=0.22 cm] (qddot) {\scriptsize $\ddot{q}$};

\end{tikzpicture}}
    \caption{The bias-variance tradeoff as a function of model expressiveness.}
    \label{fig:bvtradeoff}
\end{figure}

Both of the aforementioned approaches have limitations, summarized in \Cref{fig:bvtradeoff}. 
The assumptions made by the domain expert may not capture hard-to-model effects, leading to inaccuracies via \textit{model bias}. 
On the other hand, while the black-box approach of training an overparameterized function class may be capable of capturing the phenomena present in the training data, it often requires infeasibly large amounts of training data to achieve generalization, due to \textit{model variance}.

Ideally, we would want to take a \textit{gray-box} approach that models the parts of the system for which we believe that models are accurate, while capturing the difficult-to-model system dynamics by training a highly expressive function class such as a neural network from data.
However, typical approaches to fitting system dynamics with neural networks do not allow us to easily incorporate prior knowledge.
Although there have been proposals to learn \textit{offset functions} that correct for model bias, they are still restricted to specific use cases~\cite{ratliff2016doomed, nguyen2010using}.

In this work, we propose a structured approach to gray-box modeling of mechanical systems. 
Instead of treating the equations of motion governing the system as an arbitrary functional mapping, we make the single assumption that the system conforms to \textit{Lagrangian dynamics}. 
Consequently, we propose to learn the \textit{Lagrangian} of the system, as well as a structured representation of the forces that act on the system. 
By doing so, we can parameterize the space of all mechanical systems in a structured and modular manner. 
From these two functions, we can evaluate the accelerations acting on the system by using the \textit{Euler-Lagrange equation}. 

The method we present has the flexibility to incorporate as much prior knowledge as in a white-box approach, and, in the event of having no prior knowledge whatsoever, the method remains as expressive as a black-box approach.
However, we will show that even in this scenario, our method achieves lower model variance than naive approaches to black-box modeling owing to the constraints of physical compliance. 

By modularizing the hypothesis class into functions that represent the system's Lagrangian and functions that represent the forces acting on it, we gain the principal benefit of being able to incorporate prior knowledge where it is available. 
For example, say we only know \textit{a priori} that the system is \textit{control affine} (i.e., there exists a linear mapping between actuator inputs and the torques applied to the system) and that no forces other than gravity act on the system. 
In such a situation, we can use an expressive function class to model the system's Lagrangian, while using a much more restricted function class to model the control-affine torques, thereby reducing model variance. 

This paper is organized as follows:
\Cref{sec:relatedwork}, overviews related approaches to modeling the dynamics of mechanical systems.
\Cref{sec:background}, describes the foundations of Lagrangian dynamics, the theoretical background to gradient-based model fitting, as well as approaches to model-based control and reinforcement learning. 
Next, \Cref{sec:methodology}, describes our modular parameterization of mechanical system dynamics, as well as our methodology for fitting such models to data. 
Finally, \Cref{sec:experiments}, uses an actuated double-pendulum as a test-bed to demonstrate that our method can seamlessly incorporate prior knowledge in a flexible and expressive hypothesis class. 
Additionally, we demonstrate that using our approach over naive approaches to black-box modeling leads to improved model-based reinforcement learning. 
Source code demonstrating our work can be found at \url{https://github.com/sisl/mechamodlearn}.

\section{Related Work}
System identification has been a field of much interest to the robotics and controls community for decades. 
More recently, the techniques of \textit{inverse kinematics} and \textit{feedback linearization} have been used to actuate robots to track desired motion trajectories. 
Such approaches, like many other model-based controllers, rely on high-gain Proportional-Derivative controllers to compensate for inaccurate dynamics models, leading to non-compliant and potentially dangerous behavior~\cite{ratliff2016doomed}. 
Additionally, model-learning is of interest to the reinforcement learning community because of its data-efficiency compared to model-free methods~\cite{tu2018gap} and potential for transfer during continual learning~\cite{Laroche2017TransferRL,zhang2018decoupling}.
Hence, learning dynamics models using data, or simply fine-tuning models crafted from prior knowledge, has been approached in several ways.

White-box approaches to system identification attempt to simply calibrate the kinematic and inertial properties of an otherwise analytically specified dynamics model~\cite{an1985estimation}.
Typically starting with an analytic model or the topology of the system, the system dynamics can be cast as follows:
\begin{equation}
    \label{eqn:sysid}
    H(q,\qdot,\qddot)\cdot \theta = \tau(q,\qdot,u)
\end{equation}
Here, $q, \qdot$, and $\qddot$ are the generalized coordinates, velocities, and accelerations of the system, $u$ are the actuator inputs, and $\tau$ is the resulting generalized torque applied to the system. 
The functions $H(q,\qdot,\qddot)$ and $\tau(q,\qdot,u)$ are given by the analytic model of the system, which are related by the learnable kinematic and inertial parameters $\theta$. 
These parameters may be learned from data using least-squares regression. 

The problem with white-box approaches is that analytic specifications of many poorly understood phenomena affecting the dynamics are not always accurate, such as interactions involving contact, fluids, friction, or wear-and-tear of robotic joints. 
Simplified analytic models of such phenomena are likely to introduce \textit{model bias}, in that no set of parameters can be selected for the model that would accurately represent the phenomenon.

At the other end of the spectrum, black-box approaches to system dynamics typically use highly expressive function classes to capture the full range of phenomena in the data~\cite{werbos1992handbook, raissi2018multistep, chen1990non}. 
A neural network is commonly chosen to express the function class, taking as inputs $q, \qdot$ and $u$, and outputting the predicted acceleration, $\qddot$, as depicted in \Cref{fig:bvtradeoff}.
However, highly expressive function classes such as neural networks are typically \textit{overparameterized}. 
This means that, given a finite amount of data, many choices of parameters could perfectly fit the data. 
As a consequence, had the data used in training been slightly different, one would expect the predictions in poorly sampled parts of the input space to change dramatically.
This phenomenon is typically referred to as \textit{overfitting}, leading to prediction error due to \textit{model variance}.
Model variance is remedied by supplying large amounts of training data, though this is typically impractical in real-world robotic scenarios. 
There are black-box approaches that fit limited data using Gaussian processes~\cite{deisenroth2011pilco}, though such approaches do not scale well with dataset size or state-space dimensionality.

As summarized in \Cref{fig:bvtradeoff}, the white-box approach of relying solely on models leads to model bias, while taking the black-box approach of relying on a highly expressive function class to learn dynamics leads to model variance. 
There have been attempts to take \textit{gray-box} approaches to combining prior knowledge with the expressive function class such as a neural network, or the data-efficiency of Gaussian Processes, attempting to make the ideal compromise between bias and variance. 
One such approach is to take a gradient-based approach to learning an `offset function' that captures the discrepancy between the white-box dynamics model and the true dynamics~\cite{ratliff2016doomed}.
By using a black-box model such as a neural network to represent the offset function, they are able to capture the effects of phenomena not accounted for in the analytic model. 
Another approach learns the offset function using data-efficient Gaussian Process regression~\cite{nguyen2010using}.
The approach we present can be considered a generalization of these approaches. 
In our approach, this offset function is one of many modules available to the practitioner to plug in to the learned dynamics model in lieu of an analytic representation obtained from prior knowledge. 
Furthermore, while the aforementioned approaches~\cite{ratliff2016doomed,nguyen2010using}, focus on allowing black-box functions to model non-conservative forces, we show how we can use neural networks to also model a system's Lagrangian, which captures passive dynamics.

Another closely related work developed concurrently and independently to this work is that of Lutter, \etal~\cite{lutter2018deep}. 
Akin to our work, the authors encode the physical prior of Lagrangian mechanics into the model architecture.
However, there are key distinctions between the approaches that allow our work to serve a greater scope than that presented by Lutter, et al.~\cite{lutter2018deep}.
Firstly, they propose to directly model the conservative forces acting on the system by using a neural network. 
In contrast, we choose to model the system's potential energy and derive the conservative forces from it, thereby guaranteeing that the forces are conservative. Further, by explicitly modeling the potential energy, we are able to use variational integrators that are appropriate for contact-rich simulation.
Secondly, Lutter, et al.~\cite{lutter2018deep} make the limiting assumption that all generalized torques are directly measurable (which requires precise knowledge of the control input Jacobian) as opposed to learned functions of the system's state and input.
We instead explicitly model generalized forces in a manner flexible enough to encode prior knowledge regarding their structure, but general enough to capture complex phenomena such as dissipative friction.
These key distinctions make our methodology more generally applicable to a wider range of robotic settings.

Finally, there exists an approach for explicitly learning the topology of a mechanical manipulator~\cite{ledezma2017fopnet}. 
The work
ignores generalized forces, assuming that the generalized forces applied to the system are directly measurable.
Furthermore, it requires learning in a higher-dimensional \textit{maximal coordinate} space, while our approach directly learns on a lower dimensional constraint manifold by using \textit{minimal coordinates}.

\label{sec:relatedwork}

\section{Background}
\label{sec:background}

Our work draws inspiration from a variety of fields including classical mechanics, system identification, model-based control, and modern machine learning techniques such as deep learning:

\subsection{Lagrangian Dynamics}

The purpose of modeling a dynamic system is to be able to predict how the state of the system evolves over time. 
Formally, the state of a system is described using \textit{generalized coordinates} $q \in \bbR^N$ and velocities $\qdot \in \bbR^N$, where $N$ is the number of coordinates.
For any mechanical system, the \textit{kinetic energy} can be written as:
\begin{equation}
    T(q, \qdot) = \frac{1}{2}\qdot^T \M(q) \qdot 
\end{equation}
where $\M(q)$ is positive definite and called the \textit{mass matrix} of the system.
Furthermore, the \textit{potential energy} of the system can be defined as a scalar function $V(q)$.
Together, these energies specify the \textit{Lagrangian} of a rigid body system as:
 \begin{equation}
    \label{eqn:rbdlag}\mathcal{L}(q, \qdot) = T(q, \qdot) - V(q) 
\end{equation}

With knowledge of a system's Lagrangian, the dynamics of the system are specified by the Euler-Lagrange (EL) equation:
\begin{equation}
    \label{eqn:el}
    \ELeqn
\end{equation}
where $F(q,\qdot,u)$ represents the \textit{generalized forces} that act on the system, and $u\in\bbR^M$ are the actuator inputs. 
In the case of a control-affine system, for example, we would set $F(q,\dot{q},u) = \B(q)\cdot u$, where $\B(q) \in \bbR^{N\times M}$. 
While the EL equation implicitly specifies the system's continuous time dynamics, the explicit form of the dynamics can be found by combining \Cref{eqn:rbdlag,eqn:el}.
By applying the chain-rule, we get what is commonly referred to as the \textit{manipulator equation}~\cite{murray1994mathematical}:
\begin{equation}
    \label{eqn:manip}
    \M(q)\qddot + \C(q,\qdot)\qdot + G(q) = F(q,\qdot,u)
\end{equation}
Here, $G(q)=-\nabla_q V(q)$ are the conservative forces that act on the system, and
\begin{equation}
\label{eqn:coriolis}
    \C_{ij}(q,\dot{q}) = \frac12 \sum_{k=1}^N \left(
    \frac{\partial \M_{ij}}{\partial q_k} + 
    \frac{\partial \M_{ik}}{\partial q_j} - 
    \frac{\partial \M_{jk}}{\partial q_i}\right)\dot{q}_k 
\end{equation}
is the \textit{Coriolis matrix} of the system~\cite{murray1994mathematical}. 
Hence, if we have a function predicting the mass matrix $\M(q)$, the potential $V(q)$, and the forces $F(q,\dot{q},u)$ that act on the system, we have fully specified the continuous time dynamics of the system.\footnote{See \Cref{app:coriolis}$^2$ for a more efficient method for computing $\C(q,\qdot)\qdot$.} 

\subsection{Prediction}
Suppose we are given $\M(q)$, $V(q)$, and $F(q,\dot{q},u)$, as well as initial conditions $q_t, \qdot_t$ and actuator input $u_t$, and want to predict $q_{t'},\dot{q}_{t'}$ for some $t' = t + \Delta t$.
To make this prediction, we need to simulate the model forward in time, which requires integrating \Cref{eqn:manip}.
We first form an explicit equation for the generalized acceleration: 
\begin{equation}
\label{eqn:xdot}
\qddot = \M^{-1}(q)\left[F(q,\qdot,u) - \C(q,\qdot)\qdot  - G(q)\right]
\end{equation}
Next, we define the generalized state at time $t$ as $x_t = [q_t,\dot{q}_t]$. 
We then find $x_{t'}$ using the Runge-Kutta fourth order (RK4) integration scheme~\cite{runge1895}.
The details of how we implement RK4 can be found in~\Cref{app:rk4}.\footnote{\scriptsize See \url{http://rejuvyesh.com/publications/mechamodlearn.pdf}}

Though RK4 is considered the standard approach to fixed time-step explicit integration, there are other integration schemes. 
For example, if we have access to the Lagrangian of a system, we can use an implicit integration scheme called \textit{variational integration}, which better handles non-smooth dynamics~\cite{manchester2017variational}. 
Although we do not consider non-smooth dynamics in this work, access to these techniques is a principal benefit of explicitly modeling a system's Lagrangian.

\subsection{System Identification}
In classical system identification, \Cref{eqn:manip} is typically re-written in the linear least squares form as in~\Cref{eqn:sysid}:
\begin{equation}
    \label{eqn:lsqloss}
    \min_{\theta} \frac1N \sum_{i=1}^N \norm{H_i \theta - \tau_i}^2
\end{equation}

For smaller datasets, this can be solved to global optimality in closed form.
For larger datasets, we can take a gradient-based approach to minimize the same loss in~\Cref{eqn:lsqloss}.

Alternatively, if parameters cannot be factored into the linear representation shown in \Cref{eqn:lsqloss}, then it is common to minimize error in generalized accelerations predicted by the model~\cite{ratliff2016doomed, meier2016towards} parameterized by $\theta$:
\begin{equation}
    \label{eqn:sysidloss}
    \min_{\theta} \frac1N \sum_{i=1}^N \norm{\qddot_i - \hat{\qddot}_i(\theta)}^2
\end{equation}
where $\hat\qddot$ is found according to \Cref{eqn:xdot}.
Here, $\theta$ can correspond to the parameters of some function approximator such as a neural network.

\subsection{Model-Based Control}
\label{sec:mbc}
Given a model specifying a system's dynamics, many techniques from control theory can force a system to follow a specified trajectory, or to track some set point~\cite{danielliberzon2012}. 
Since we limit the scope of experiments in this work to smooth dynamical systems, we use Direct Collocation Trajectory Optimization (DIRCOL)~\cite{kelly2017trajopt} in order to generate a `nominal trajectory' ($\bar{x}_{1:t_H}, \bar{u}_{1:t_H}$) that:
\begin{enumerate}
    \item Transports the system from some initial condition to some desired set point,
    \item Is dynamically feasible according to the model, and
    \item Minimizes some cost function defined over the trajectory, which is typically quadratic in the control effort and tracking error.
\end{enumerate}
To track the nominal trajectory in real-time, we use a Time-Varying Linear Quadratic Regulator (TVLQR)~\cite{briananderson2007}. Here, the system dynamics are linearized along the nominal trajectory from DIRCOL, and feedback gains $K_t$ are found using dynamic programming to minimize a quadratic cost function.
The actuator input to the system at some time $t$ is:
\begin{equation}
    u_t = \pi(x_t) = \bar{u}_t - K_t (x_t - \bar{x}_t)
\end{equation}
where $\pi:\bbR^N\to \bbR^M$ is referred to as the \textit{synthesized policy}.

\subsection{Model-Based Reinforcement Learning}
\label{sec:mbrl}
Assume we have a finite-horizon Markov decision process $\mathcal{M} = \{\mathcal{X},\mathcal{U},\model, R, H\}$, where $\mathcal{X}$ is the state-space, $\mathcal{U}$ is the action-space, $\model:\mathcal{X}\times\mathcal{U}\rightarrow\mathcal{X}$ is the system dynamics-model, $R:\mathcal{X}\times\mathcal{U}\to\bbR$ is the reward function, and $H$ is the horizon-length. 
In our context, $\mathcal{X} \in \bbR^{2N}$ is the space of all generalized coordinates and velocities, and $\mathcal{U} \in \bbR^M$ is the space of all actuator inputs. 
Additionally, we assume that while $\model(x,u)$ is not known \textit{a priori}, the reward function $R(x,u)$ is known. 

As discussed in \Cref{sec:mbc}, if $\model(x,u)$ were known, we could use DIRCOL to synthesize a trajectory that optimally solves our task, and a \textit{policy} $\pi:\mathcal{X}\to\mathcal{U}$ that tracks this trajectory using TVLQR. 
If $\model(x,u)$ is not known, it can be learned from data observed while interacting with the environment. 

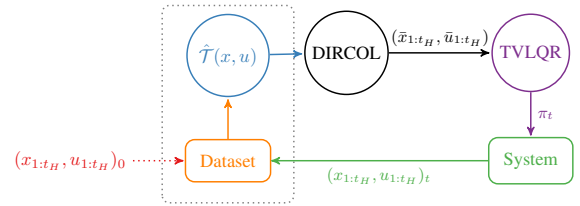
\begin{figure}
    \centering
    \scalebox{0.7}{\begin{tikzpicture}[->, >=stealth', shorten >=1pt, auto, node distance=0.8cm, thick, main/.style={draw, inner sep=8pt, outer sep=0pt, rounded corners=5pt}]
    \node[circle, draw=nice-blue, inner sep=4pt] (model) at (1.25, -1.00) {\color{nice-blue} $\hat{\model}(x, u)$};
    \node[]  (tau_1) at (-1.75, -3.00) {\color{nice-red}  $({x_{1:t_H}, u_{1:t_H}})_0$};
    \node[circle, draw=black, inner sep=3pt] (synth) at (3.50, -0.95) {DIRCOL};
    \node[circle, draw=C4, inner sep=3pt] (track) at (7.00, -0.95) {\color{C4} TVLQR};
    \node[main, draw=nice-orange] (data) at (1.25, -3.00) {\color{nice-orange}  Dataset};
    \node[main, draw=nice-green] (env) at (7.00, -3.0) {\color{nice-green}  System};
    \node [inner sep=0pt] (hid) at (2.00, -3.00) {};
    \draw[dotted, draw=gray, rounded corners] (0.0, -3.8) rectangle ++(2.50, 3.75) {};
    \draw[dotted, nice-red] (tau_1) to (data);
    \draw[nice-orange] (data) to node [auto] {} (model);
    \draw[nice-blue] (model) to (synth);
    \draw[] (synth) to node [auto] {\small $(\bar{x}_{1:t_H}, \bar{u}_{1:t_H})$} (track);
    \draw[C4] (track) to node [auto] {\small $\pi_t$} (env);
    \draw[nice-green] (env) to node [auto] {\small \color{nice-green} $(x_{1:t_H}, u_{1:t_H})_t$} (hid);
\end{tikzpicture}}
    \caption{Framework for Model-Based Reinforcement Learning}
    \label{fig:mbrl_framework}
\end{figure}

\Cref{fig:mbrl_framework} shows the architecture for model-based reinforcement learning used in this work.
Polydoros \etal provide a survey of a variety of other approaches~\cite{Polydoros2017}.
Initially, an episode of training data is generated by randomly actuating the system and observing its state-transitions. 
A model, $\hat{\model}(x,u)$ is learned from the dataset. 
The model is then passed through DIRCOL and TVLQR to generate a policy $\pi(x)$ that, if the model were accurate, would solve the task. 
Naturally, the model being imperfect, this would not solve the task, but visit novel states. 

By repeating this process, novel data is added to the dataset, allowing the learned model to approach the true model in accuracy. 
Once its prediction accuracy is close enough to the true model, the synthesized policy $\pi(x)$ will be able to solve the task on the real system. 
However, it is possible for this approach to fail due to a lack of \textit{exploration}. 
That is, if we attempt to \textit{greedily} solve the task with our model at every episode, the policies being followed may never visit states that are required in the dataset in order to learn a sufficiently accurate model. 

A simple approach to exploration in this setup is to add \textit{exploration noise} that adds variance to the policies generated by DIRCOL+TVLQR. 
Here, we add random perturbations to the nominal trajectory synthesized by DIRCOL before passing it to TVLQR. 
By doing so, the system is encouraged to visit states in the neighborhood of what is otherwise believed to be an optimal trajectory.

\section{Methodology}
\label{sec:methodology}

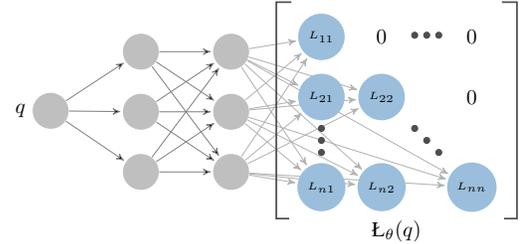
\begin{figure}
    \centering
    \scalebox{0.8}{\begin{tikzpicture}[shorten >=1pt,->,draw=black!50]

    \def\vertdots[#1,#2]{
        \draw[black!70,fill=black!70] (#1,#2+0.2) circle (0.05 cm);
        \draw[black!70,fill=black!70] (#1,#2) circle (0.05 cm);
        \draw[black!70,fill=black!70] (#1,#2-0.2) circle (0.05 cm);

    }

    \def\hordots[#1,#2]{
        \draw[black!70,fill=black!70] (#1+0.2,#2) circle (0.05 cm);
        \draw[black!70,fill=black!70] (#1,#2) circle (0.05 cm);
        \draw[black!70,fill=black!70] (#1-0.2,#2) circle (0.05 cm);

    }

    \def\slantdots[#1,#2]{
        \draw[black!70,fill=black!70] (#1+0.2,#2-0.2) circle (0.05 cm);
        \draw[black!70,fill=black!70] (#1,#2) circle (0.05 cm);
        \draw[black!70,fill=black!70] (#1-0.2,#2+0.2) circle (0.05 cm);

    }

    \tikzstyle{neuron}=[circle,fill=gray!50,minimum size=17pt] %
    \tikzstyle{bigneuron}=[circle,fill=nice-blue!50, minimum size=22pt] %

    \def\layersep{1.5 cm}
    \def\nodesep{1 cm}

    \pgfmathsetmacro{\ilw}{1}
    \pgfmathsetmacro{\nhl}{2}
    \pgfmathsetmacro{\hlw}{3}
    \pgfmathsetmacro{\N}{3}

    \pgfmathtruncatemacro{\iof}{(\ilw*\nodesep)/2}
    \pgfmathtruncatemacro{\hof}{(\hlw*\nodesep)/2}
    \pgfmathtruncatemacro{\oof}{(\N*\nodesep+0.5*\nodesep)/2}

    \foreach \name / \y in {1,...,\ilw}
        \path[yshift=\iof]
            node[neuron] (I-\name) at (0,-\y * \nodesep) {};

    \foreach \namei / \yi in {1,...,\nhl}
        \foreach \namej / \yj in {1,...,\hlw}
            \path[yshift=\hof]
                node[neuron] (H-\namei\namej) at (\yi * \layersep,-\yj * \nodesep) {};

    \foreach \name / \y in {1,...,\N}{
        \foreach \nname / \yt in {1,...,\y}{

            \ifnum\y<\N
                \ifnum\yt<\N
                    \path[yshift=\oof]
                        node[bigneuron] (O-\name\nname) at (\nhl*\layersep + \layersep + \yt*\nodesep - \nodesep,-\y * \nodesep) {\tiny $L_{\y\yt}$};
                \else
                    \path[yshift=\oof]
                    node[bigneuron] (O-\name\nname) at (\nhl*\layersep + \layersep + \yt*\nodesep - \nodesep/2,-\y * \nodesep) {\tiny $L_{\y n}$};
                \fi
            \else
                \ifnum\yt<\N
                    \path[yshift=\oof]
                        node[bigneuron] (O-\name\nname) at (\nhl*\layersep + \layersep + \yt*\nodesep - \nodesep,-\y * \nodesep-\nodesep/2) {\tiny $L_{n\yt}$};
                \else
                    \path[yshift=\oof]
                    node[bigneuron] (O-\name\nname) at (\nhl*\layersep + \layersep + \yt*\nodesep - \nodesep/2,-\y * \nodesep-\nodesep/2) {\tiny $L_{nn}$};
                \fi
            \fi
            };
        };

    \vertdots[\nhl*\layersep + \layersep, -1]
    \hordots[6.25,0.75]
    \slantdots[6.25,-1]

    \pgfmathtruncatemacro{\Nmo}{\N-1}
    \foreach \name / \y in {1,...,\Nmo}{
        \foreach \nname / \yt in {1,...,\y}{
            \ifnum\y>1
                \ifnum\yt>1
                    \path[yshift=\oof]
                        node at (\nhl*\layersep + \layersep + \N*\nodesep - \yt*\nodesep,-\N * \nodesep + \y * \nodesep) {$0$};
                \else
                    \path[yshift=\oof]
                    node at (\nhl*\layersep + \layersep + \N*\nodesep - \yt*\nodesep + \nodesep/2,-\N * \nodesep + \y * \nodesep) {$0$};
                \fi
            \else
                \ifnum\yt>1
                    \path[yshift=\oof]
                        node at (\nhl*\layersep + \layersep + \N*\nodesep - \yt*\nodesep,-\N * \nodesep + \y * \nodesep) {$0$};
                \else
                    \path[yshift=\oof]
                    node at (\nhl*\layersep + \layersep + \N*\nodesep - \yt*\nodesep + \nodesep/2,-\N * \nodesep + \y * \nodesep) {$0$};
                \fi
            \fi
            };

        };

    \foreach \source in {1,...,\ilw}
        \foreach \dest in {1,...,\hlw}
            \path (I-\source) edge (H-1\dest);

    \ifthenelse{\nhl > 1}{
        \pgfmathtruncatemacro{\nhlm}{\nhl - 1}
        \foreach \layer in {1,...,\nhlm}
            \pgfmathtruncatemacro{\nextlayer}{\layer + 1}
            \foreach \source in {1,...,\hlw}
                \foreach \dest in {1,...,\hlw}
                    \path (H-\layer\source) edge (H-\nextlayer\dest);
        }{}

    \begin{scope}[on background layer]
    \foreach \source in {1,...,\hlw}
        \foreach \desta / \ya in {1,...,\N}
            \foreach \destb / \yb in {1,...,\ya}
                \path [black!30] (H-\nhl\source) edge (O-\desta\destb);
    \end{scope}

    \draw[thick, black!70, -] (4,1.3) -- (3.75,1.3) -- (3.75,-2.3) -- (4,-2.3);
    \draw[thick, black!70, -] (4+3.5,1.3) -- (4.25+3.5,1.3) -- (4.25+3.5,-2.3) -- (4+3.5,-2.3);

    \node[left of = I-1, xshift=0.5 cm] (q) {$q$};

    \node (Lth) at (5.75 cm,-2.5 cm) {$\L_\theta(q)$};

\end{tikzpicture}}
    \caption{A neural network parameterization for the Cholesky factor of the model's inertia matrix.}
    \label{fig:Mnet}
\end{figure}

In its most general form, we model a physical system by modeling:
\begin{enumerate}
    \item Its Lagrangian, and
    \item The generalized forces that act on it.
\end{enumerate}

We first present a methodology for modeling the positive-definite mass-matrix $\M_\theta(q)$ using a neural network with parameters contained in $\theta$. 
As shown in \Cref{fig:Mnet}, we predict $\M_\theta(q)\succ 0$ by predicting the $\frac{N^2+N}{2}$ elements of its Cholesky factor, which is a lower-triangular matrix $\L_\theta(q)$, as is done by~\cite{lutter2018deep,haarnoja2016backprop}.\footnote{One can optionally enforce that the diagonal elements of $\L_\theta(q)$ are positive in order to make the matrix the \textit{unique} Cholesky factor of $\M_\theta(q)$, though we have empirically found this to be unnecessary and only makes model parameters' optimization more difficult.}
Here, the neural network first outputs a vector, of which the first $N$ elements are used as the diagonal of $\L_\theta(q)$, and the remaining $\frac{N^2-N}{2}$ are used for the off-diagonal elements of the lower half of the matrix. 
We additionally add a constant offset to the diagonals of $\L_\theta(q)$ so that $\M_\theta(q)$ is diagonally dominant and easily invertible given random initializations of $\theta$. 
We then predict:
\begin{equation}
    \M_\theta(q) = \L_\theta(q)\L^\top_\theta(q)
\end{equation}
Additionally, we predict the potential energy, $V_\theta(q)$, using a neural network that maps $q\in\bbR^N\to \bbR$, and in the most general case, the generalized forces as a function $F_\theta(q,\qdot, u)$ that maps $\bbR^{2N+M}\to \bbR^N$. 
Since we are required to take gradients of $\M_\theta(q)$ as well as $V_\theta(q)$, with restpect to $q$, in order to compute the system's acceleration, and occasionally Hessians of the Lagrangian in order to measure properties of the model such as local generalized stiffness, we require the non-linearities in the neural network to be at least twice-differentiable. 
In this work, we use the hyperbolic tangent function (tanh) as the non-linearity.

We have thus far presented the parameterization for the most generic form of a mechanical system. 
If any prior knowledge is available to a practitioner, they may substitute $\M_\theta(q)$, $V_\theta(q)$, or $F_\theta(q,\qdot,u)$ with a more restricted function class than a neural network, so long as as $\M_\theta(q)$ and $V_\theta(q)$ remain twice-differentiable. 
For example, one may wish to model the system as control-affine with viscous joint-damping. 
This can be achieved by substituting:
\begin{equation}
    F_\theta(q,\qdot,u) = \B_\theta(q)\cdot u + \eta_\theta(q) \circ \qdot
\end{equation}
Here, $\B_\theta(q) \in \bbR^{N\times M}$ and $\eta_\theta(q) \in \bbR^N$. 
Though this parameterization is still fairly expressive, it is possible that there are phenomena in the data that it is unable to accurately model.

If $\M_\theta(q)$, $V(q)$ and $F(q,\qdot,u)$ are all substituted with analytic models derived from prior knowledge, then this model reduces to a white-box model. 
On the other hand, if left in its most general form, then the model is still as expressive as a black-box model. 
This is because $\M_\theta(q)$ is capable of expressing a function always equal to the identity matrix, and $V(q)$ is capable of expressing a function always equal to zero. 
If this were the case, then it can easily be seen from \Cref{eqn:xdot} that $F_\theta(q,\qdot,u)$ reduces to the black-box neural network model show in \Cref{fig:bvtradeoff}.
However, we will show empirically that despite having such expressive power, we still gain a reduction in model variance over a naive black-box model.

\subsection{Parameter Optimization}
Having described the parameterization for a model of an arbitrary mechanical system, we now describe how we optimize these parameters to learn a model of a system from data.
Given some dataset of a system's state-transitions:
\begin{equation}
    \mathcal{D}=\left\{q_k,\qdot_k,u_k,q'_k,\qdot'_k \mid k\in\{1,\ldots N\}\right\}
\end{equation}
We optimize the parameters $\theta$ by minimizing the \textit{prediction loss}:
\begin{equation}
    \label{eqn:ourloss}
    L(\theta) = \frac1N \sum_{k=1}^N \norm{q'_k - \hat{q}'_k(\theta)}^2 + \lambda \norm{\qdot'_k - \hat{\qdot}'_k(\theta)}^2
\end{equation}
where $\hat{q}'_k, \hat{\qdot}'_k$ are predicted using RK4 applied to the model dynamics with inputs $q_k, \qdot_k, u_k$ and parameters $\theta$. 
We then minimize this loss using gradient descent methods such as Adam~\cite{kingma2014adam}. 
We use the loss presented in \Cref{eqn:ourloss} as opposed to that presented in \Cref{eqn:sysidloss} for two reasons. 
Firstly, accelerations $\qddot_k$ are often not directly measured, and rather estimated by finite-difference approximation in practice, which amplifies high-frequency noise present in the data samples. 
By opting for the loss we presented, we do not need to estimate $\qddot$ as the regression target. 
Secondly, we have observed that minimizing prediction error is a better metric to optimize a model against than aligning accelerations predicted by the model with those estimated in the data. 
The down-side of our approach, however, is that differentiating through RK4 in order to compute parameter gradients is a constant-factor more expensive than computing gradients for the loss presented in \Cref{eqn:sysidloss}.

\section{Experiments}
\label{sec:experiments}

In our experiments, we aim to justify the following properties of our method:
\begin{itemize}
    \item Ability to learn the dynamics of a complex mechanical system from data,
    \item Seamless incorporation of prior knowledge to reduce model variance,
    \item Better data-efficiency for model learning than a naive, black-box approach even in the absence of prior knowledge, and,
    \item More data-efficient model-based reinforcement learning. %
\end{itemize}

We do so by performing three experiments. 
In the first, we vary the amount of prior knowledge used when modeling the system and analyze the effect on data required for the learned model to generalize accurately, i.e. analyze the reduction on model variance gained by using prior knowledge.
In the second experiment, we compare the ability of a naive approach with our structured black-box approach to make accurate long-horizon predictions.
In the third experiment, we perform model-based reinforcement-learning and compare our method with the use of naive-black box model to parameterize the modeled dynamics.

\subsection{Experimental Domain}
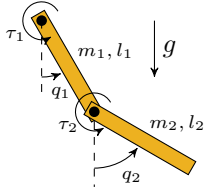
\begin{figure}
    \centering
    \scalebox{1.0}{\begin{tikzpicture}

	\draw[fill=C3, rotate around={30:(0,0)}] (-0.1,0.1) rectangle (0.1,-1.5);

	\fill (0,0) circle [radius=2pt];
	\node (a) at (-60:1.4) {};

	\draw[fill=C3, rotate around={60:(a)}] ($(a)+(-0.1,0.1)$) rectangle ($(a)+(0.1,-1.5)$);

	\fill (a) circle [radius=2pt];

	\node (r1cm) at (-60:0.7) {};
	\node (r2cm) at ($(a) + (-30:0.7)$) {};

	\node () at ($(r1cm) + (0.5,0.2)$) {\scriptsize$m_1, l_1$};

	\node () at ($(r2cm) + (0.5,0.2)$) {\scriptsize$m_2, l_2$};

	\draw[dashed] (0,0) -- (0,-1.0);
	\draw[->] (-90:0.75) arc (-90:-70:0.75);
	\node () at (-75:1) {\scriptsize $q_1$};

	\draw[dashed] (a) -- ($(a) + (0,-1.0)$);
	\draw[->] ($(a)+(-90:0.75)$) arc (-90:-40:0.75);
	\node () at ($(a)+(-60:1)$) {\scriptsize $q_2$};

	\draw[->] ($(0,0)+(30:0.25)$) arc (30:300:0.25);

	\node () at ($(0,0) + (210:0.4)$) {\scriptsize$\tau_1$};

	\draw[->] ($(a)+(30:0.25)$) arc (30:300:0.25);

	\node () at ($(a) + (210:0.4)$) {\scriptsize$\tau_2$};

	\draw[->] (1.5,0) -- (1.5,-0.75);

	\node () at (1.7,-0.375) {$g$};

\end{tikzpicture}}
    \caption{The actuated double pendulum system.}
    \label{fig:double_pen}
\end{figure}

We first focus on the fully actuated double pendulum (shown in \Cref{fig:double_pen}) as our system of interest, which has highly non-linear dynamics that are chaotic when unforced. 
As shown in the figure, the system is specified by the masses $m_1$ and $m_2$ and lengths $l_1$ and $l_2$ of the rods, the gravitational acceleration $g$, and coefficients that specify the joint torques $\tau_1$ and $\tau_2$, which will be introduced shortly. 
The system has two generalized coordinates, $q_1 \in [-\pi,\pi)$ and $q_2\in[-\pi,\pi)$, which correspond to the relative deflection of each rod in radians, as well as control inputs $u_1$ and $u_2$. 
The dynamics of the system are computed by specifying the mass-matrix, $\M_{sys}(q)$, the potential energy, $V_{sys}(q)$, and the generalized coordinates. 
The specifications for $\M_{sys}(q)$ and $V_{sys}(q)$ can be found in~\Cref{app:exp}.
The generalized forces acting on the system are composed of a control torque, as well as viscous joint-damping:
\begin{equation}
    F_{sys}(q,\qdot,u) = \begin{bmatrix}\tau_1\\ \tau_2\end{bmatrix} = \begin{bmatrix}b_1 & 0 \\ 0 & b_2\end{bmatrix} \begin{bmatrix}u_1\\u_2\end{bmatrix} + \begin{bmatrix}\eta_1\\\eta_2\end{bmatrix}\circ \begin{bmatrix}\qdot_1\\\qdot_2\end{bmatrix}
\end{equation}
Here, $b_1$ and $b_2$ are the coefficients of the control-matrix $\B_{sys}(q)$, and $\eta_{sys}$ specify the linear joint-damping coefficients. 
Hence, the nine white-box free parameters $\theta_{wb}$ of this model are:
\begin{equation}
    \theta_{wb} = \left[m_1,m_2,l_1,l_2,g,b_1,b_2,\eta_1,\eta_2 \right]
\end{equation}

\subsection{Data-efficiency and Prior Knowledge}
\label{sec:dateff}
In this experiment, we aim to justify the claim that prior knowledge in the form of analytic specifications for $\M(q)$, $V(q)$, $\B(q)$, or $\eta$ can be easily incorporated into the model parameterization and thereby improve the data-efficiency of learning. 
We compare with a naive black-box approach in which $\qddot = NN(q,\qdot,u;\theta)$, where $NN(\cdot;\theta):\bbR^{2N + M}\rightarrow \bbR^N$ is a feedforward neural network.

The model in which $\M_\theta(q)$, $V_\theta(q)$, $F_\theta(q,\qdot,u)$ are all represented by neural networks in the manner introduced in \Cref{sec:methodology} is as expressive as the naive model. 
Here, no prior knowledge is encoded aside from the assumption that the system conforms to Lagrangian dynamics.
In addition to comparing these two models, we compare models where some components are represented by neural networks, corresponding to a lack of prior knowledge, while others are represented by the same analytic functions as that of the true system, except that the parameters of those functions are learned from data. 
\Cref{table:mdlpnl} lists the models compared. 
Components with the subscript $\theta$ are represented in neural networks, and those with the subscript $wb$ are ones represented by analytic functions.
Specifically, the trainable parameters in $\Mwb, \Vwb, \B_{wb}$ and $\eta_{wb}$ are $\{\hat{m}_1,\hat{m}_2,\hat{l}_1,\hat{l}_2\}$, $\{\hat{m}_1,\hat{m}_2,\hat{l}_1,\hat{l}_2, \hat{g}\}$, $\{\hat{b}_1, \hat{b}_2\}$, and $\{\hat{\eta}_1,\hat{\eta}_2\}$, respectively. 

Since the white-box parameterization, referred to as {`W-B'}, is exactly that of the true system, we would expect it to be able to perfectly represent the dynamics of the true system with only a handful of data points. 
On the other hand, since the naive approach incorporates no prior knowledge whatsoever, we would expect it to require the largest amount of training data to avoid model variance and accurately generalize learned dynamics to unseen states. 
Additionally, we would expect all other compared modules to require less data the more prior knowledge they incorporate. 

To test this hypothesis, we sample a dataset $\{q,\qdot,u,q',\qdot'\}\in\Dtr$ and $\{q,\qdot,u,q',\qdot'\}\in\Dte$ where $q\sim\mathcal{U}(-\pi,\pi)~[\si{\radian}]$, $\qdot\sim\mathcal{U}(-10,10)~[\si{\radian\per\second}]$, $u\sim\mathcal{N}(0,120^2)$, and $(q',\qdot')$ are found by passing the system dynamics and the sampled state and input through RK4 with a time-step of $\Delta t = 0.01~\si{\second}$. 

We first estimate the amount of training data, i.e. $\lvert \Dtr\rvert$, needed in order for the naive model to make accurate predictions on $\Dte$. 
In order to do so, we begin with $\lvert\Dtr\rvert = 8$, doubling $\lvert\Dtr\rvert$ until we find the loss on $\Dtr$ after training to completion is similar to the loss on $\Dte$, implying that the naive model is generalizing well. 
This procedure gives a range, $\{\lvert\Dtr\rvert_\text{min},\lvert\Dtr\rvert_\text{max}\}=\{2^{12},2^{13}\}$, which contains the minimum $\lvert\Dtr\rvert$ required for the naive model to generalize. 
When given a training dataset of $2^{13}$ samples, the naive model achieves a validation loss according to \Cref{eqn:ourloss} of $10^{-2.5}$. 

Starting with $\lvert\Dtr\rvert = 8$, we then train each of the models in \Cref{table:mdlpnl}, checking if a training dataset of that size is sufficient for the model to achieve a validation loss of $10^{-2.5}$. 
If it is not, we double the dataset size and repeat the process. 

\begin{figure}[t]
    \centering
    \scalebox{0.7}{\begin{tikzpicture}[]
\begin{axis}[ylabel = {Minimum Dataset Size}, ymode = {log}, xlabel = {Modules with NN Parametrizations}, log basis y={2}, stack plots=y, /pgf/bar width=3pt, x tick label style={rotate=70}, xtick = {1,2,3,4,5,6,7,8,9,10,11}, xticklabels = {W-B, B, F, V, M, MB, VB, MV, MVB, MVF, NaiveNN}]\addplot+ [mark = {none}, ybar,draw=none, fill=none]coordinates {
(1, 8)
(2, 64)
(3, 1024)
(4, 1024)
(5, 256)
(6, 256)
(7, 2048)
(8, 1024)
(9, 1024)
(10, 2048)
(11, 4096)
};
\addplot+ [mark = {none}, ybar, draw=C1, fill=C1!50]coordinates {
(1, 8)
(2, 64)
(3, 1024)
(4, 1024)
(5, 256)
(6, 256)
(7, 2048)
(8, 1024)
(9, 1024)
(10, 2048)
(11, 4096)
};
\end{axis}

\end{tikzpicture}}
    \caption{Minimum amount of training data required for various models to generalize well across the Actuated Double Pendulum state-space.}
    \label{fig:data_complexity}
\end{figure}
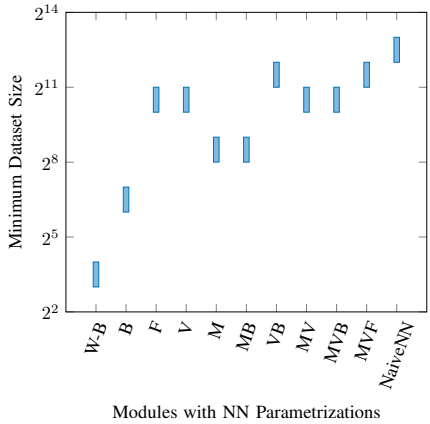

This procedure ultimately gives us a range, for each model, specifying the minimum amount of data required in order for that model to achieve the stated validation loss. 
\Cref{fig:data_complexity} shows the range found for each model in \Cref{table:mdlpnl}. 
The experiment was run for 5 different seeds that vary the initializations of trainable parameters, as well as the samples making up the train and validation sets. 
We find the results to be consistent across seeds. 

Examining \Cref{fig:data_complexity} from left to right, we see that as we reduce the number of components parameterized with prior knowledge, the amount of data required for the model to generalize well, i.e. for the model to avoid suffering from model variance, increases. 
We note that the white-box parameterization, which has $9$ trainable parameters, takes between 8 and 16 data points to train, matching expectations. 
Further, we note that the MVF parameterization, which incorporates no prior knowledge except that the system conforms to Lagrangian dynamics, requires on the order of half as much data to generalize as well as the naive approach does. 
The B parameterization requires on the order of 16 times less data to generalize compared to the F parameterization. 
This is not particularly surprising as the fully generally parameterization of the forcing function used in F is far more expressive than the control-affine parameterization used in B. 
We also observe that having no prior knowledge regarding the potential energy, i.e. letting $V(q)$ be represented by a neural network, increases the data required to generalize, suggesting that one may want to consider less expressive models for it. 

In summary, this experiment shows that the methodology we present allows us to learn dynamics of a complex mechanical system from data, and seamlessly incorporate available prior knowledge by substituting neural network parameterizations for the various components with less expressive parameterizations. 
Additionally, we show that even in the absence of prior knowledge, i.e. the MVF parameterization, using our approach enables us to generalize effectively from less data than a naive black-box approach.

\begin{table}[t]
\centering
 \begin{tabular}{r | r r r} 
 \toprule
 Name & $\M$ prm & $V$ prm & $F$ prm \\ [0.5ex] 
 \midrule
 MVF & $\Mnn$ & $\Vnn$ & $\Fnn$ \\ 
 MVB & $\Mnn$ & $\Vnn$ & $\Bnn$ \\
 MV & $\Mnn$ & $\Vnn$ & $\Fwb$ \\
 VB & $\Mwb$ & $\Vnn$ & $\Bnn$ \\
 MB & $\Mnn$ & $\Vwb$ & $\Bnn$ \\
 M & $\Mnn$ & $\Vwb$ & $\Fwb$ \\
 V & $\Mwb$ & $\Vnn$ & $\Fwb$ \\
 B & $\Mwb$ & $\Vwb$ & $\Bnn$ \\
 F & $\Mwb$ & $\Vwb$ & $\Fnn$ \\
 W-B & $\Mwb$ & $\Vwb$ & $\Fwb$ \\ [1ex]
 \bottomrule
\end{tabular}
\caption{Various models incorporating varying degrees of prior knowledge. `prm' is used as shorthand for `parameterization'.}
\label{table:mdlpnl}
\end{table}

\subsection{Multi-Step Prediction on the Double Pendulum}
\begin{figure}[t]
    \centering
    \scalebox{0.475}{\input{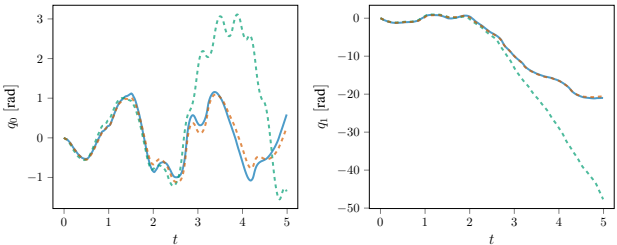}}
    \caption{MVF ({\color{C2}{orange}}) vs Naive ({\color{nice-green}{green}}) $5$\si{\second} predictions with the same initial conditions and inputs as the true trajectory ({\color{C1}{blue}}).}
    \label{fig:tspreds}
\end{figure}
In this experiment, we compare the ability of the Naive and MVF models to perform multi-step predictions on the double-pendulum domain.
Both models are trained on the same dataset of 4,096 samples. 
\Cref{fig:tspreds} shows that the MVF model is able to accurately replicate the behavior of the true system for the entirety of the 5 second simulation, while the naive model quickly diverges from ground truth. 
This result is consistent with the conclusion that the MVF model more accurately learns the system's dynamics from limited data when compared to the naive model. 
Experiments thus far have trained the models on i.i.d data sampled from an \textit{a priori} specified distribution over the input-space. 
In the next experiment, we demonstrate the ability to efficiently learn dynamics in a reinforcement learning setting, in which data is presented in the form of trajectories sampled from the system, and thereby not i.i.d.

\subsection{Model-based reinforcement learning}
In this experiment, we compare the naive model with the MVF and MVB models from~\Cref{table:mdlpnl} in their abilities to learn the dynamics of the actuated double pendulum for the purpose of solving a swing up task. 
The task is to actuate the system in a manner that brings it to an state that is upright and stationary, i.e. $(q_1,q_2) = (\pi, 0.0)~[\si{\radian}]$ and $(\qdot_1,\qdot_2) = (0.0,0.0)~[\si{\radian\per\second}]$, in 2.06 seconds, and then remain stable in this configuration for an additional 0.5 seconds. 
In order to do so, the model must learn from the available data well enough to accurately generalize predicted dynamics to the relevant parts of the state space.
If the model is inaccurate, we expect DIRCOL to produce nominal trajectories that are less dynamically feasible, and TVLQR to produce less accurate linearizations of the dynamics about those trajectories. 
These two effects will consequently cause the algorithm to require many more interactions with the environment in order to solve the task, when using that model to plan a trajectory. 
Hence, we expect to see that the naive model requires the largest number of interactions to solve the task, followed by MVF, and then MVB. 

We follow the procedure for model-based reinforcement learning described in \Cref{sec:mbrl}. 
All trajectories start from the downward equilibrium $(q_1, q_2) = (0.0, 0.0)$ and $(\qdot_1, \qdot_2) = (0.0, 0.0)$.
We provide all models the same initial trajectory found by randomly actuating the system with actions $u_t\sim\mathcal{N}(0,120^2)$. 
All trajectories are simulated with $\Delta t = 0.1~\si{\second}$, and where the action $u_t$ applied is clipped to have a maximum absolute value of $120$. 
We then train all models on this dataset for 5000 epochs, using the Adam optimizer with learning rate of $3\times10^{-4}$. 
Every time new data is added to a model's dataset, the model is trained for an additional 1000 epochs.

Next, we perform trajectory optimization using the trained models to get nominal trajectories and a TVLQR tracking controller to synthesize a policy for each model. 
However, in order to encourage exploration, we add noise sampled from $\mathcal{N}(0,0.25^2)$ to the nominal trajectories independently to each $q_{1,t},q_{2,t},\qdot_{1,t},\qdot_{2,t}$ and $u_t$. 
Each model's policy is then rolled out on the system, and the trajectory generated is added to each model's dataset. 
This process is repeated until the trajectories followed by using a given model are reliably solving the task. 
We repeat this test using three different random seeds.

We measure the performance of the system by measuring the mean Euclidean distance between the end effector and its desired set-point (vertical and stable), over the trajectory. 
A well-performing model will have a low score using this metric. 
When evaluating performance, we do not add noise to the nominal trajectories.

\begin{figure}[t]
    \centering
     \scalebox{0.7}{\input{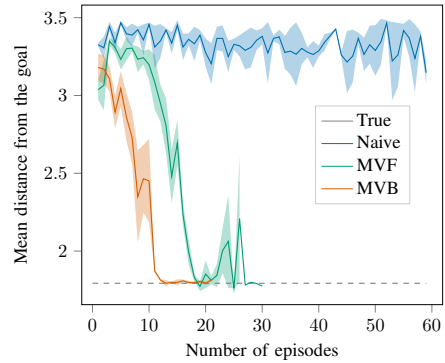}}
    \caption{Model-based reinforcement learning performance of various model parameterizations. Here, performance is the mean Euclidean distance between the end-effector and the target over the trajectory, where a lower score is better.}
    \label{fig:mbrl}
\end{figure}

\Cref{fig:mbrl} shows the results of this experiment. 
As we can see, the results match our expectations: the MVB parameterization, which incorporates some prior knowledge about the structure of the generalized forces, solves the task with the fewest number of interactions with the environment. 
Additionally, the MVF parameterization, which incorporates no prior knowledge, takes more interactions to solve the task, but does so eventually. 
The naive black-box parameterization does not succeed in solving the task in the allowed time-frame. 
Observing the nominal trajectories and the policies' abilities to follow the trajectories on the real system, we see the behavior we expect for MVF and MVB.\footnote{Videos of the training episodes can be found at \url{https://youtu.be/NxHVLlNj6hg}.}
The nominal trajectories eventually start looking more realistic, and the policies start doing a better job of following them. 
However, when using the naive parameterization, DIRCOL appears to be unable to find a solution that transports the system to the desired goal state. 
Consequently, the trajectories added to the dataset never explore the relevant parts of state-space, resulting in a model that is never able to generalize well enough to solve the task.

Seeing that even the MVF model, which incorporates no prior knowledge, is able to solve the task in a reasonable amount of time, as well as the fact that prior knowledge improves sample efficiency, validates the hypothesis that using the method we presented enables more efficient model-based reinforcement learning than if a naive black-box parameterization of the dynamics were used.

\section{Conclusions}
\label{sec:conclusion}

In this work we presented a method for parameterizing an arbitrary mechanical system using neural networks, as well as a method for training such models from data. 
Unlike naive black-box approaches that predict accelerations directly, we use neural networks to parameterize the Lagrangian of a system and the generalized forces that act on it. 
We showed that such a modular parameterization is flexible enough to allow us to seamlessly incorporate prior knowledge where it is available, allowing a practitioner to precisely balance model bias and variance. 
We showed on a simulated actuated double pendulum
that, even in the absence of prior knowledge, our method learns the dynamics of complex mechanical systems more efficiently than a naive, black-box approach. 

There are many interesting applications of this work that we intend to explore in the near future. 
Firstly, by explicitly modeling the system's Lagrangian, we have the ability to use variational integrators to accurately simulate non-smooth dynamics. 
By interfacing optimal control strategies with such simulators, we can perform model-based reinforcement learning in contact-rich environments. 
Secondly, by parameterizing the space of mechanical systems, we can take scalable probabilistic approaches to system-identification. 
This allows us to explore possibilities for robust control and safe, efficient model-based reinforcement learning. 
Lastly, a limitation of our current framework is that it is more computationally expensive to make predictions with our model than with a black or white-box model.
We intend to explore optimizations of our method that can close this performance gap.

\section*{Acknowledgments}
We are thankful to Jeannette Bohg for advice. 
This work is supported in part by DARPA under agreement number D17AP00032.
The content is solely the responsibility of the authors and does not necessarily represent the official views of DARPA.

\bibliographystyle{IEEEtran}
\bibliography{references}

\clearpage
\onecolumn
\appendices
\crefalias{section}{appsec}
\section{Efficiently Computing Coriolis Forces}
\label{app:coriolis}
In this section, we describe a method for computing $\C(q,\qdot)\qdot$ in $\mathcal{O}(N^2)$. We begin by restating \Cref{eqn:coriolis}:
\begin{equation}
\label{eqn:coriolis2}
    \C_{ij}(q,\dot{q}) = \frac12 \sum_{k=1}^N \left(
    \frac{\partial \M_{ij}}{\partial q_k} + 
    \frac{\partial \M_{ik}}{\partial q_j} - 
    \frac{\partial \M_{kj}}{\partial q_i}\right)\dot{q}_k 
\end{equation}

We can find the the $i^{th}$ element of $C(q,\qdot)\qdot$ as follows:
\newcommand\dijdk{\ensuremath{\frac{\partial \M_{ij}}{\partial q_k}}}
\newcommand\dikdj{\ensuremath{\frac{\partial \M_{ik}}{\partial q_j}}}
\newcommand\dkjdi{\ensuremath{\frac{\partial \M_{kj}}{\partial q_i}}}
\newcommand\sumj{\ensuremath{\sum_{j=1}^N}}
\newcommand\sumk{\ensuremath{\sum_{k=1}^N}}
\begin{equation}
\begin{aligned}
    \{\C(q,\qdot)\qdot\}_i &= \sumj \left(\frac12 \sumk \left(
    \dijdk + 
    \dikdj - 
    \dkjdi \right)\dot{q}_k\right) \qdot_j \\ 
    &= \frac12 \underbrace{\sumj\sumk \dijdk\qdot_k\qdot_j}_{\text{S}1} + \frac12\underbrace{ \sumj\sumk \dikdj\qdot_k\qdot_j}_{\text{S}2} - \frac12 \sumj\sumk \dkjdi\qdot_k\qdot_j 
\end{aligned}
\end{equation}
Note the symmetry between $\text{S}1$ and $\text{S}2$. 
Combining and moving gradients outside the summation, we get:
\begin{equation}
\begin{aligned}
    \{\C(q,\qdot)\qdot\}_i  
    &= \sumj \frac{\partial }{\partial q_j}\sumk \M_{ik}\qdot_k\qdot_j - \frac{\partial }{\partial q_i} \sumj\sumk \frac12\M_{kj}\qdot_k\qdot_j 
\end{aligned}
\end{equation}
Which can be concisely written as:
\begin{equation}
    \C(q,\qdot)\qdot = \nabla_q\left(\M(q) \qdot\right)\qdot - \nabla_q\left(\frac12\qdot^\top\M(q)\qdot\right)
\end{equation}
The most expensive operation is the computation of the Jacobian matrix $\nabla_q(\M(q)\qdot)$, which is computed in $\mathcal{O}(N^2)$.

\section{Prediction with RK4}
\label{app:rk4}
Given a the system's generalized coordinates $q_t$ and generalized velocity $\qdot_t$ at some time $t$, we wish to predict $q_{t'}$ and $\qdot_{t'}$ at some time $t' = t + \Delta t$. First, we define $x_t = [q_t,\qdot_t]^\top$, and consequently, $\dot{x}_t = [\qdot_t,\qddot_t]^\top$. Here, $\qddot_t$ is computed according to \Cref{eqn:xdot}. 
To integrate a function $g(x, u)$, RK4 is a standard integration scheme:
\begin{equation}
    \begin{aligned}
        k_1 &= \Delta t \cdot g(x_t, u_t) \\
        k_2 &= \Delta t \cdot g(x_t + k_1/2, u_t) \\
        k_3 &= \Delta t \cdot g(x_t + k_2/2, u_t) \\ 
        k_4 &= \Delta t \cdot g(x_t + k_3, u_t) \\
        x_{t'} &= x_t + \frac16 (k_1 + 2k_2 + 2k_3 + k_4) \\
    \end{aligned}
\end{equation}
Here, we let $g([q,\qdot]^\top,u) = [\qdot,\qddot(q,\qdot,u)]^\top $.

\section{Experiments}
\label{app:exp}
Here we specify the double pendulum dynamics in our experiments, as well as parametrizations of the naive black-box model and structured black box models used in our experiments.
\subsection{Double Pendulum}
The mass matrix $\M_{sys}(q)$ and potential energy $V_{sys}(q)$ of the double pendulum are analytically specified below. 
\begin{equation}
    \M_{sys}(q) = \begin{bmatrix} 
        I_{11} & I_{12} \\
        I_{12} & I_2
    \end{bmatrix}
\end{equation}
where
\begin{align}
    I_1 &= \frac{1}{3} m_1 l_1^2 \\
    I_2 &= \frac{1}{3} m_2 l_2^2 \\
    I_{11} &= I_1 + I_2 + m_2 {l_1}^2 + m_2 l_1 l_2 \cos{q_2} \\
    I_{12} &= I_2 + \frac{1}{2} m_2 l_1 l_2 \cos{q_2}
\end{align}
\begin{equation}
    V_{sys}(q) = - \frac{1}{2} m_1 g l_1 \cos{q_1} - m_2g\left(l_1 \cos{q_1} + \frac{l_2}{2}\cos(q_1+q_2)\right)
\end{equation}

For the experiments, the parameters specified in \Cref{tab:sysparams} are used to define the system dynamics.
\begin{table}[h]
\centering 
\begin{tabular}{ccs|ccs} \toprule
    {Parameter} & {Value} & {Unit}&{Parameter} & {Value} & {Unit} \\ \midrule
    $m_1$  & 10.0 & \si{\kilogram} & $b_1$  & 1.0 & \si{\newton\meter} \\  
    $m_2$  & 10.0 & \si{\kilogram} & $b_2$  & 1.0 & \si{\newton\meter} \\
    $l_1$  & 1.0 & \si{\meter} & $\eta_1$  & -0.5 & \si{\newton\meter\second\per\radian} \\
    $l_2$  & 1.0 & \si{\meter} & $\eta_2$  & -0.5 & \si{\newton\meter\second\per\radian} \\
    $g$  & 10.0 & \si{\meter\per\second^2} & & & \\ \bottomrule
\end{tabular}
\caption{Parameters specifying the system dynamics for both experiments.}
\label{tab:sysparams}
\end{table}

\subsection{Model Parameterization}
The Naive model consists of a multi-layer perceptron with 3 hidden layers of dimension 64.
The MVF model consists of a multi-layer perceptron for $\M(q)$, $V(q)$ and $F(q, \qdot, u)$ each, with 3 hidden layers of dimension 32. 
These network sizes are chosen so that the naive and MVF model have roughly the same number of trainable parameters. 
All other models listed in \Cref{table:mdlpnl} that use neural networks to parameterize components use neural networks with with 3 hidden layers of dimension 32 for those components.

\end{document}